\def\year{2021}\relax
\documentclass[letterpaper]{article} 
\usepackage{aaai21}  
\usepackage{times}  
\usepackage{helvet} 
\usepackage{courier}  
\usepackage[hyphens]{url}  
\usepackage{graphicx} 
\usepackage{natbib}  
\usepackage{caption} 
\usepackage{bm}
\usepackage{array}
\usepackage{multirow}
\usepackage{amsmath}
\usepackage{amsfonts}
\usepackage{graphicx}
\usepackage{color}
\usepackage{arydshln}
\usepackage{tikz}

\usepackage[noend]{algpseudocode}
\usepackage{tikz}
\usepackage{booktabs}
\usepackage{tikzsymbols}
\usepackage[linesnumbered, ruled, vlined]{algorithm2e}
\algnewcommand\RETURN{\algorithmicreturn}
\newcommand*{\Scale}[2][4]{\scalebox{#1}{$#2$}}  
\DeclareMathOperator*{\argmax}{arg\,max}
\DeclareMathOperator*{\argmin}{arg\,min}

\title{C2C-GenDA: Cluster-to-Cluster Generation for Data Augmentation of Slot Filling}
\author{
	Yutai Hou\footnotemark[1],
	Sanyuan Chen\footnotemark[1],
	Wanxiang Che\footnotemark[2],
	Cheng Chen,
	Ting Liu
	\\
	\normalfont{Research Center for Social Computing and Information Retrieval, Harbin Institute of Technology, China} \\
	{\tt \{ythou, sychen, car, tliu\}@ir.hit.edu.cn,} {\tt 170400202@stu.hit.edu.cn} \\
}

\begin{document}
	
	\maketitle
	\renewcommand{\thefootnote}{\fnsymbol{footnote}}
	\footnotetext[1]{Equal contributions.}
	\footnotetext[2]{Corresponding author.}
	\renewcommand{\thefootnote}{\arabic{footnote}}

\begin{abstract}
Slot filling, a fundamental module of spoken language understanding, often suffers from insufficient quantity and diversity of training data.
To remedy this, we propose a novel Cluster-to-Cluster generation framework for Data Augmentation (DA), named C2C-GenDA.
It enlarges the training set by reconstructing existing utterances into alternative expressions while keeping semantic.
Different from previous DA works that reconstruct utterances one by one independently, C2C-GenDA jointly encodes multiple existing utterances of the same semantics and simultaneously decodes multiple unseen expressions.
Jointly generating multiple new utterances allows to consider the relations between generated instances and encourages diversity.
Besides, encoding multiple existing utterances endows C2C with a wider view of existing expressions, helping to reduce generation that duplicates existing data.
Experiments on ATIS and Snips datasets show that instances augmented by C2C-GenDA improve slot filling by 7.99 (11.9\%$\uparrow$) and 5.76 (13.6\%$\uparrow$) F-scores respectively, when there are only hundreds of training utterances.\footnote{Code: \url{https://github.com/Sanyuan-Chen/C2C-DA}}

\end{abstract}

\section{Introduction}
\begin{figure}[t]
	\centering
	\footnotesize
	\begin{tikzpicture}
	\draw (0,0 ) node[inner sep=0] {\includegraphics[width=0.85\columnwidth, trim={9.5cm 4.5cm 9.7cm 2.8cm}, clip]{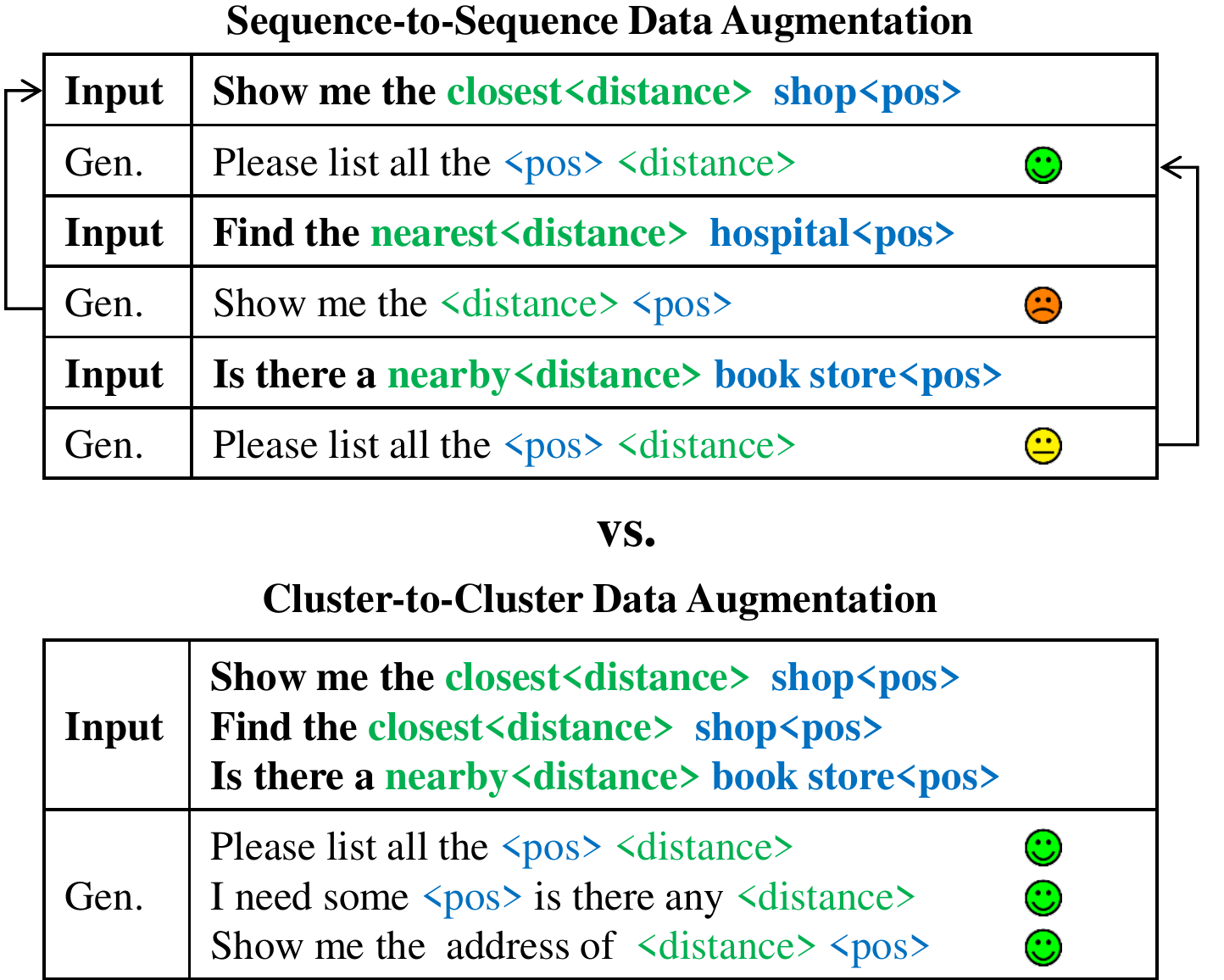}};
	\end{tikzpicture}
	\caption{\footnotesize
		Examples of sequence-to-sequence data augmentation and cluster-to-cluster data augmentation. \cChangey{2} denotes novel utterance. \cChangey{-1} denotes duplication to existing utterance. \cChangey{0} denotes duplication to other generated utterances.
	}\label{fig:intro}
\end{figure}

Slot filling is a fundamental module of the Spoken Language Understanding (SLU) in the task-oriented dialogue system \cite{young2013pomdp}. 
The ``inputs'' in Figure \ref{fig:intro} shows examples of slot filling, where key entities within user utterances are tagged with slot labels. 
Due to the high cost of manual annotation and the rapidly changing nature of dialogue domain, slot filling often faces the lack of quantity and diversity of training data.
Such insufficiency in training data poses serious challenges for slot-filling models to handle myriad ways in which users express their demands.

\textit{Data augmentation} (DA) technique, which improves diversity and quantity of training data with synthetic instances, 
offers an appealing solution to the data scarcity problem of SLU.
Success has been achieved with data augmentation on a wide range of problems, including computer vision \cite{NIPS2012_4824}, speech recognition \cite{DBLP:journals/corr/HannunCCCDEPSSCN14}, 
text classification \cite{zhang2015character}, and question answering \cite{fader-zettlemoyer-etzioni:2013:ACL2013}.

For slot filling, state-of-the-art data augmentation works focus on generative methods \cite{shin2019vae}.
One of their typical ideas is generating new utterances by reconstructing existing utterances into alternative expressions while keeping the semantics.
Previous works learn a Sequence-to-Sequence (Seq2Seq) model to reconstruct each existing utterance one-by-one \cite{yoo2020deep,hou2018coling,DBLP:conf/interspeech/KurataXZ16}.
However, these methods tend to generate duplicated utterances, because they can only consider the expression variance between one input-output pair at a time.
For example in Figure \ref{fig:intro}, each new utterance is only generated to be different from the corresponding input utterance, and thus often unconsciously duplicates other generated utterances (\cChangey{0}) or other input utterances (\cChangey{-1}). 
Such duplication will hinder the effectiveness of data augmentation.
We argue these defects can be easily avoided by breaking the shackles of current one-by-one augmentation paradigm and considering the extensive instance relations during generation.

In this paper, we propose a novel Cluster-to-Cluster Generation framework for Data Augmentation of slot filling, named C2C-GenDA.
As shown in Figure \ref{fig:intro}, different from previous works that augment each utterance one-by-one independently, we jointly generate multiple new instances by reconstructing a cluster of existing utterances with the same semantics. 
Such cluster-to-cluster generation allows model to consider the duplication between generated utterances and aware of more existing expressions in original data. 
These advantages of C2C-GenDA remedy the aforementioned defects of Seq2Seq DA and help to improve generation diversity. 
To encourage diversity and quality of generation, we propose the Duplication-aware Attention and Diverse-Oriented Regularization mechanisms,
both of which promote diverse decoding.
To learn to generate diverse new utterances, we train the C2C-GenDA model with cluster-to-cluster `paraphrasing' pairs, and introduce a Dispersed Cluster Pairing algorithm to extract these cluster pairs from existing data. 

Experiments on ATIS and Snips datasets show that the proposed method significantly improves the performance of slot-filling systems. 
Case studies and analysis of augmented data also confirm that our method generates diverse utterances. 
Our contributions can be summarized as follow
(1) We propose a novel Cluster-to-Cluster generation framework for data augmentation of slot filling, which can remedy the duplication problem of existing one-by-one generation methods.
(2) We propose the Duplication-aware Attention and Diverse-Oriented Regularization mechanism to improve diversity of the augmented utterances.
(3) We introduce a Dispersed Cluster Pairing algorithm to extract cluster-to-cluster `paraphrasing' pairs for data augmentation model training.

\section{Problem Description}

In this paper, we study the data augmentation for slot filling task that maps utterances into semantic frames (slot type and slot value pairs).
Slot filling is commonly treated as a sequence labeling problem, 
where \textit{slot type} labels are assigned to contiguous sequences of words indicating these sequences are the corresponding \textit{slot values}.

We specify the data augmentation (DA) for slot filling as 
exploiting existing training instances to generate new expressions 
for each semantic frame.
Suppose existing slot filling training data is $D=\{(\mathbf{u}_{i}, \mathbf{s}_{i})\}_{i=1}^N$.
Given a semantic frame $\mathbf{s}_j$ and the corresponding existing utterances $C=\{\mathbf{u} | (\mathbf{u}, \mathbf{s}) \in D \wedge \mathbf{s}=\mathbf{s}_j\}$, 
DA generates a set of new utterances $C'=\{\mathbf{u'}_{i}\}_{i=1}^{M'}$ with unseen expressions.
Then DA constructs new training instances $D'_{\mathbf{s}_j}=\{(\mathbf{u'}_{i}, \mathbf{s}_j)\}_{i=1}^{M'}$ by associating new utterances with the semantic frame. 
Finally, DA takes the union of all new instances $D'=\bigcup_{\mathbf{s}_j} D'_{\mathbf{s}_j}$ as the additional data to reinforce the model training.
\section{Proposed Framework}
In this section, we present an overview of our data augmentation framework, 
and introduce the Cluster2Cluster generation model.
Then, we discuss how to extract cluster-to-cluster paraphrasing data for generation model training.

\subsection{Overview}\label{sec:over}
Here, we introduce the overview of the proposed cluster-to-cluster data augmentation framework for slot filling.
For each semantic frame, 
we use a Cluster2Cluster (C2C) model to generate new expressions from existing utterances.
The input of our framework is a cluster of existing instances for a certain semantic frame, 
and the output is a cluster of generated new instances with unseen expressions.

Following \citet{hou2018coling}, we perform delexicalized generation.
Specifically, both the inputs and outputs of C2C generation model are delexicalized utterances, where slot values tokens are replaced by slot label tokens.
For the example in the Figure \ref{fig:intro}, C2C takes in ``show me the $<$distance$>$ $<$pos$>$'' and reconstruct the expression as ``please list all the $<$pos$>$ $<$distance$>$ ''.
The delexicalization focuses the model on generating diverse expressions rather than slot values and reduces the vocabulary size.
Then after generation, we recover the delexicalized utterances by filling the slots with context-suitable slot values.
Such delexicalization is important since it allow us to generate both the utterance and accurate slot annotations simultaneously.

To learn the ability of generating diverse and new expressions, 
we construct cluster-to-cluster paraphrasing pairs from original training data with the Dispersed Cluster Pairing algorithm, 
which simulates the data augmentation process of generating novel expressions from existing expressions for a specific semantic frame.

\begin{figure*}[t]
	\centering
	\footnotesize
	\begin{tikzpicture}
	\draw (0,0 ) node[inner sep=0] {\includegraphics[width=1.9\columnwidth, trim={0cm 2.8cm 1.2cm 5.2cm}, clip]{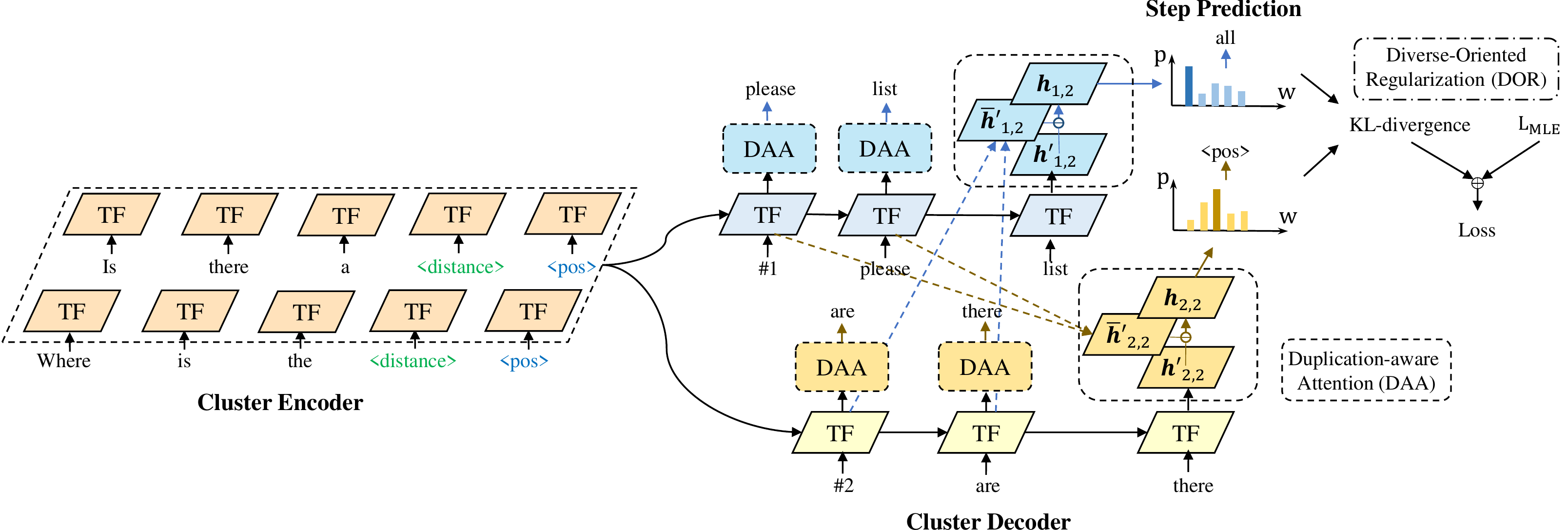}};
	\end{tikzpicture}
	\caption{\footnotesize
		Cluster2Cluster generation model
	}\label{fig:C2C}
	\vspace*{-2mm}
\end{figure*}

\subsection{Cluster2Cluster Generation Model}\label{sec:c2c}
Custer2Cluster (C2C) model is a generation model that lies at the core of our C2C-GenDA framework and aims to reconstruct input utterances into alternative expressions while keeping semantic.
As Figure \ref{fig:C2C} shown, the C2C model first encodes the input cluster of utterances $C = \{\mathbf{u}_i \}^M_{i=1}$ for a certain semantic frame, then jointly decodes a new cluster of utterances  $C' = \{ \mathbf{u}'_i \}^{M'}_{i=1}$ with different expressions, where $M$ and $M'$ are size of input and output cluster respectively. 

To further encourage the diversity of the generated utterances, we propose two novel mechanisms:
(1) \textbf{Duplication-aware Attention} that attends to the existing expressions to avoid duplicated generation for each decoding step.
(2) \textbf{Diverse-Oriented Regularization} that guides the synchronized decoding of multiple utterances to improve the internal diversity of the generated cluster. 

%

\paragraph{Cluster Encoder}
We jointly encode multiple input utterances by concatenating them, and representing the whole sequence with an $L$-layer transformer \cite{Transformer}:\footnote{We separate input utterances with special tokens $<$SEP$>$.}
\[
\Scale[0.85]{
	\begin{array}{ll}
	\vspace*{1mm}
	\bm{H}_0 & = \{\bm{e}_{i,1}, \bm{e}_{i,2}, ..., \bm{e}_{i, n_i}\}_i^{M} \\
	\vspace*{1mm}
	\bm{A}_l & = {\rm LN}(\bm{H}_{l-1} + {\rm MultiHead}(\bm{H}_{l-1}, \bm{H}_{l-1})) \\
	\vspace*{1mm}
	\bm{H}_l & = {\rm LN}(\bm{A}_l + {\rm FFN}(\bm{A}_l)) \\
	\vspace*{1mm}
	\bm{R} & = \bm{H}_L \\
	\end{array}
}
\]
where $\bm{R}$ is the final representations of the input tokens, $\bm{e}_{i,j}$ is the embedding of $j_\text{th}$ token in the $i_\text{th}$ input utterance, $\bm{H}_0$ is the package of all input token embeddings, and $\bm{H}_l$ is the outputs of $l_\text{th}$ layer.
${\rm LN}$ is the layer normalization. ${\rm MultiHead}(\bm{Q}, \bm{V})$ is the multi-head self-attention function operating on vector packages of queries $\bm{Q}$, values $\bm{V}$ (also used as keys). ${\rm FFN}$ is a position-wise feed-forward network.

\paragraph{Cluster Decoder with Duplication-aware Attention}
To reduce duplication and encourage diversity, we propose a cluster decoder with the Duplication-aware Attention (DAA) mechanisms.
It decodes each new utterance while being aware of the existing expressions in both the input cluster and other generated utterances.

Intuitively, we decode the $r_{th}$ target utterance $\mathbf{u'}_r$ depending on both input cluster $C$ and other output utterances $C' \setminus \{\mathbf{u'}_r\}$.
We also incorporate the diversity rank token \#$r$ \cite{hou2018coling} as generation conditions to encourage diversity and distinguish different output utterances. 
Details of the diverse rank in C2C will be introduced in a later section. 
Then the C2C model is formalized as:
\[
\Scale[0.85]{
	p(C' \mid C) = p(\mathbf{u'}_1, ..., \mathbf{u'}_{M'}  \mid C) = \displaystyle \prod_r^{M'}{p(\mathbf{u'}_r \mid C, \text{\#}r, C' \setminus \{\mathbf{u'}_r\}}). \\
}
\]

However, it is unrealistic to decode a target utterance depending on all the other target utterances, because we jointly decode all the target utterances and the generation of other target utterances has not finished. 
Therefore, we approximate the dependence between target utterances and depend the decoding on already generated tokens of all the target utterances.
For each step, we simultaneously decode one token for all the target utterances which depends on all the previously decoded tokens $\{\mathbf{u}'_{i,1:t-1}\}_i^{M'}$: 
\[
\Scale[0.85]{
	\begin{array}{rl}
	p(C' \mid C) = \displaystyle \prod_{t}^{T}{\prod_{r}^{M'}{p(u'_{r, t} \mid C, \text{\#}r, \{\mathbf{u}'_{i,1:t-1}\}_i^{M'} )}}
	\end{array}
},
\]
where $T$ is the number of decoding steps. 

We calculate the decoding possibility for the $t_{th}$ step of $r_{th}$ utterance as $p_{r,t} = {{\rm Softmax}({\rm MLP}(\mathbf{h}_{r, t}))}$, where $\mathbf{h}_{r, t}$ is a hidden state that combines feature representations of $C$, $\text{\#}r$ and $\{\mathbf{u}'_{i,1:t-1}\}_i^{M'}$.

Here, we obtain the hidden state $\mathbf{h}_{r, t}$ with DAA which contains two terms: $\mathbf{h'}_{r, t}$ and $\mathbf{\bar{h}'}_{r, t}$.
The first term $\mathbf{h'}_{r, t}$ mainly records the information of what token should be generated. 
To achieve this, $\mathbf{h'}_{r, t}$ encodes previously decoded tokens of current utterance and semantic information from the input cluster.
Since $\mathbf{h'}_{r, t}$ encodes existing expressions in the input cluster, it also allows to reduce generation duplicated to existing expressions. 
For the $r_\text{th}$ target utterance, we compute $\mathbf{h'}_{r, t}=\bm{R}'_{r,t}$ with an $L$-layer transformer as decoder:
\[
\Scale[0.85]{
	\begin{array}{ll}
	\vspace*{1mm}
	\bm{H}'_0&=\{\#r, \bm{e}'_{r,1}, \bm{e}'_{r,2}, ..., \bm{e}'_{r, t-1}\} \\
	\vspace*{1mm}
	\bm{A}'_l&={\rm LN}(\bm{H}'_{l-1} + {\rm MultiHead}(\bm{H}'_{l-1}, \bm{H}'_{l-1} \cup \bm{H}_{l-1})) \\
	\vspace*{1mm}
	\bm{H}'_l&={\rm LN}(\bm{A}'_l + {\rm FFN}(\bm{A}'_l)) \\
	\bm{R}'_r&=\bm{H}'_L \\
	\end{array}
}
\]
where $\bm{R}'_r$ is a package of hidden states for all $t$ decoding steps. 
$\bm{H}'_0$ is the package of all decoded token embeddings and $\bm{H}'_l$ is the outputs of $l_\text{th}$ decoding layer.
$\bm{H}_l$ is the input cluster representation from $l_\text{th}$ encoding layer.

The second term $\mathbf{\bar{h}'}_{r, t}$ mainly records the duplicated expressions that should not be generated, it encodes expressions generated by other target utterances as $\mathbf{\bar{h}'}_{r, t} = {\rm MultiHead}(\mathbf{h'}_{r, t}, \{\mathbf{h'}_{i,1:t-1}\}_{i \neq r}^{M'})$. 

Finally, the hidden-state for decoding is
$\mathbf{h}_{r, t} = \mathbf{h'}_{r, t} - \lambda \cdot \mathbf{\bar{h}'}_{r, t}$, where $\lambda$ is a balance factor.
Subtraction makes $\mathbf{h}_{r, t}$ different for each target utterances, and $-\mathbf{\bar{h}'}_{r, t}$ can implicitly punish decoding of commonly shared words.

\begin{figure}[t]
	\centering
	\footnotesize
	\begin{tikzpicture}
	\draw (0,0 ) node[inner sep=0] {\includegraphics[width=0.9\columnwidth, trim={19.2cm 3.5cm 0cm 0cm}, clip]{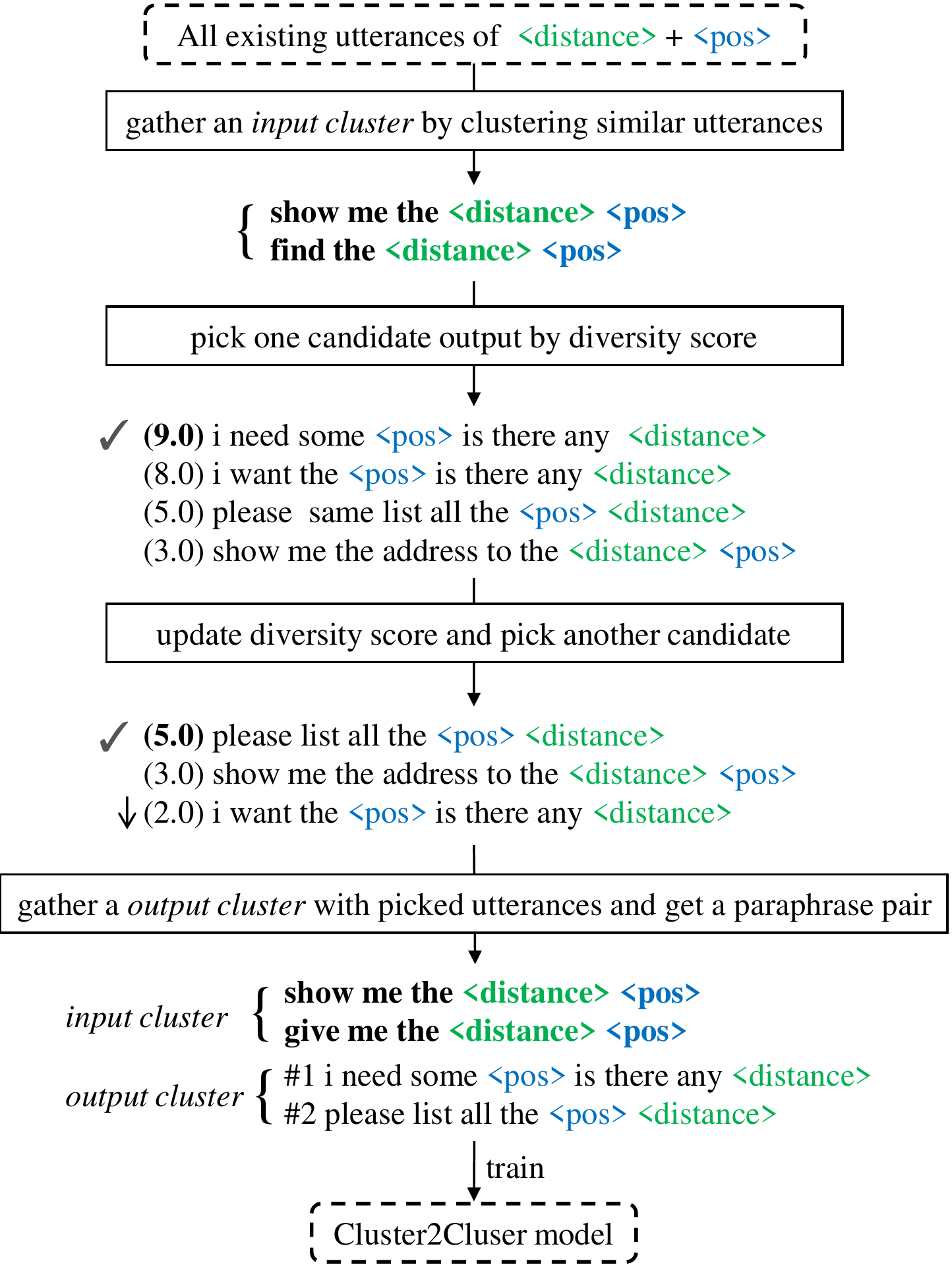}};
	\end{tikzpicture}
	\caption{\footnotesize
		Construct training instance for our Cluster2Cluster model with Dispersed Cluster Pairing.
	}\label{fig:train}
\end{figure}

\paragraph{Model Training with Diverse-Oriented Regularization}
We train the C2C model with a Diverse-Oriented Regularization (DOR) to encourage internal diversity within the generated utterance cluster.

To achieve this, we propose to enlarge the distance between distributions of utterances in the output cluster. 
However, the distribution of an utterance is hard to estimate during the decoding process. 
Thus, we approximately enlarge two utterances' distribution by encouraging the divergence of token distributions.
As shown in Figure \ref{fig:C2C}, we train the model to enlarge the Kullback-Leibler Divergence (KL) between decoding distribution of different output utterances at each step.
Formally, we define the distance between two output utterances $\mathbf{u'}_{r_i}$ and $\mathbf{u'}_{r_j}$ as:
\[
\Scale[0.85]{
	{\rm Dist}(\mathbf{u'}_{r_i}, \mathbf{u'}_{r_j}) = \sum_t{{\rm KL}(p_{{r_i}, t} || p_{{r_j}, t})}
},
\]
where $p_{{r}, t}$ denotes token distribution of ${r_{th}}$ output utterance at $t_{th}$ decoding step. 
Then we define Diverse-Oriented Regularization of generation as:
\[
\Scale[0.85]{
	R_{\rm diverse} = -\sum_{{r_i},{r_j} ({r_i} \neq {r_j})}{{\rm Dist}(\mathbf{u'}_{r_i}, \mathbf{u'}_{r_j})}
}.
\]

Overall, we train C2C model to minimize: 
\[
\Scale[0.85]{
	{Loss} = -log(p(C' \mid C)) + \gamma \cdot R_{\rm diverse}
},
\]
where $\gamma$ is a balancing factor.

%
%
%

\paragraph{Generation Pre-training}\label{sec:pretrain}
When training data is insufficient, the data augmentation model itself is often poorly trained due to limited expression in the training data. 
To remedy this, we initialize the transformer encoder/decoder with pre-trained language model GPT-2 \cite{radford2019language}.

\begin{algorithm}[t]
	\caption{Dispersed  Cluster Pairing}\label{alg:data}
	\scriptsize
	\footnotesize
	\begin{algorithmic}[1]
		\Require Original training data $D=\{(\mathbf{u}_{i}, \mathbf{s}_{i})\}_{i=1}^N$ \\
		Initialize cluster-to-cluster pairs $\mathcal{P}=\left\{ \right\}$ \\
		All semantic frame $\mathcal{S} = \{\mathbf{s} \mid \exists (\mathbf{u}, \mathbf{s}) \in D \}$ \\
		Delexicalize all training data $D' = {\rm Delexicalize}(D)$ \\
		\For{$\mathbf{s}$ in $\mathcal{S}$} {
			$C_{\mathbf{s}}=\{(\mathbf{\tilde{u}},\mathbf{\tilde{s}}) \mid (\mathbf{\tilde{u}}, \mathbf{\tilde{s}}) \in D \land \mathbf{\tilde{s}} = \mathbf{s}\}$ 
			
			$\mathcal{C}_\text{lexical}$ = ${\rm K\_Medoids}(C_{\mathbf{s}})$
			
			\For{$C=\{(\mathbf{u}_{i}, \mathbf{s}_{i})\}_{i=1}^M$ in $\mathcal{C}_\text{lexical}$} {
				
				Initialize target cluster $C'=\left\{ \right\}$
				
				\While{$|C'| < M' $ }{
					
					$\mathbf{u}'=  \displaystyle \argmax_{\mathbf{u} \in C_s \setminus C }{{\rm DS}(\mathbf{u}, C \cup C' )}$ 
					
					Update $ C' = \{\mathbf{u}'\} \cup C'$ 
				}
				Update $\mathcal{P} = \mathcal{P} \cup \{ (C, C') \} $ 
			}
		} \\
		\RETURN   $ ~\mathcal{P}$
	\end{algorithmic}
\end{algorithm}

\subsection{Cluster-to-Cluster Data Construction}\label{sec:data}
To learn to generate diverse new utterances, we train the C2C model with cluster-to-cluster `paraphrasing' pairs extracted from existing training data, and propose a \textbf{Dispersed Cluster Pairing} algorithm to construct these pairs.

We hope the cluster-to-cluster generation pairs simulate the data augmentation process, where we generate diverse new utterances from limited expressions. 
Therefore, given all utterances with same semantic, we gather similar utterances as an input cluster and pick the utterances with the most different expressions as the output cluster. 
For each semantic frame $\mathbf{s}$, we construct the input cluster $C$ with \textit{lexical clustering} and the construct output cluster $C'$ with \textit{furthest including} mechanism.

Figure \ref{fig:train} and Algorithm \ref{alg:data} present the workflow of the cluster-to-cluster data construction. 
Firstly, we perform \textit{lexical clustering} on the utterances with K-Medoids clustering method \cite{FastKM}. 
Each lexical cluster contains similar utterances and is used as an input cluster $C$. 

Then, for each source cluster $C$, we sample target utterance according to a \textit{furthest including} principle. 
Each time, we pick the $\mathbf{u'}$ that has the highest \textit{diversity score} and include it in target cluster $C'$.

We compute \textit{diversity score} between a candidate utterance $\mathbf{u}'$ and the union of source cluster $C$ and current target cluster $C'$ as ${\rm DS} = \min_{\mathbf{u} \in C \cup C' }{\textsc{EditDistance}(\mathbf{u}, \mathbf{u'})}$.
Notice that maximizing the diverse score between $\mathbf{u}'$ and source cluster $C$ increases the target cluster's novelty against the source cluster. The diverse score between $\mathbf{u}'$ and target cluster $C'$ helps to avoid duplication within the target cluster.

\paragraph{Diversity Rank}
As mentioned in the decoder section, we adopt the diversity rank to encourage diversity and distinguish sentences in the output cluster.
Consequently, we incorporate the diverse rank in training data of C2C model by associating each output utterance with a diverse rank token $\#r$ (See the examples in Figure \ref{fig:train}).
Since the output cluster utterances are greedily picked by diversity score, we naturally use this greedy picking order as the diversity rank, which models the novelty of output utterance.
When augmenting new data, we generate the new utterances at rank from 1 to $M'$, where $M'$ is a preset size of output cluster.


\paragraph{Cross Expansion}\label{sec:cross}
After training of the C2C model, we generate unseen new utterances from the constructed input clusters.
To avoid the new utterances to overfit to the original output utterances seen in C2C training, we perform data augmentation with a \textit{Cross Expansion} mechanism.
We partition all the cluster-to-cluster pairs $\mathcal{P}$ into training ones $\mathcal{P}_\text{train}$ and reserved ones $\mathcal{P}_\text{seed}$. 
Then we train the C2C model only with $\mathcal{P}_\text{train}$ and generate new utterances from the input clusters of reserved pairs $\mathcal{P}_\text{seed}$.
To make full use of existing utterances, we repeat such partition in a crossing manner. 



\begin{table*}[t]
	\centering
	\footnotesize
	\renewcommand\arraystretch{1.1}
	\begin{tabular}{lcccccc}
		\toprule
		\multirow{2}{*}{\textbf{Model}} &
		\multicolumn{3}{c}{\textbf{ATIS}} &
		\multicolumn{3}{c}{\textbf{Snips}} 	\\
		\cmidrule(lr){2-4}
		\cmidrule(lr){5-7}
		& {\textbf{Full}} & \multicolumn{1}{c}{\textbf{Medium}} & {\textbf{Small}} & {\textbf{Full}} & {\textbf{Medium}} & {\textbf{Small}} \\	
		\midrule
		Baseline & { 94.93 } & {85.85} & { 67.33 }
		& { 89.30 } & { 64.84 } & { 42.33 } \\ 
		~ + NoiseSeq2Seq \cite{DBLP:conf/interspeech/KurataXZ16} & { 94.61 } & { 87.34 } & { 67.93 }  
		& { - } & { - } & { - } \\
		~ + Slot Expansion \cite{shin2019vae} & { 94.67 } & { 87.58 } & { 74.83 } 
		& { - } & { - } & { - } \\
		~ + Rel-Seq2Seq \cite{hou2018coling} & { 94.82 } & { 88.72 } & { 73.71 }
		& { - } & { - } & { - } \\
		~ + C-VAE \cite{shin2019vae} & { 95.04 } & { 88.82 } & { 71.97 }
		& { 90.93 } & { 65.13 } & { 38.46 } \\
		\midrule
		~ + Ours  w/o pre-train  & { \textbf{95.06}} & { \textbf{90.87$^*$}} & { \textbf{75.21$^*$}} 
		&  { 90.33 } & { \textbf{67.49$^*$}} & { \textbf{46.94$^*$}} \\	
		~ + Ours  & { \textbf{95.29}} & { \textbf{90.95$^*$}} & { \textbf{75.32$^*$}} 
		&  { \textbf{91.01} } & { \textbf{67.90$^*$}} & { \textbf{48.09$^*$}} \\	
		
		\bottomrule
	\end{tabular}
	\caption{
		Comparison of data augmentation methods for slot filling on ATIS and Snips datasets.
		Results marked with \textbf{+} are results of the same Bi-LSTM trained with different data augmentation methods. 
		\textit{w/o pre-train} initializes C2C with random parameters rather than pretrained GPT.
		The Snips results are re-implemented.  \textbf{*} indicates that the result is statistically significant over the strongest data augmentation baseline under t-test (p-value $<$ 0.05). 
	}\label{tbl:main}
\end{table*}

\section{Experiment}
We evaluate the proposed data augmentation method on two slot filling datasets.\footnote{
We only focus on DA for the sequence-labeling problem of slot-filling. So the results may be lower than some joint SLU models, which perform slot-filling using additional information from intents \cite{peng2020gpt,louvan2020simple}.} 


\paragraph{Data}
We conduct experiments on ATIS and Snips datasets. 
ATIS \cite{atis} is extensively used for slot filling and provides a well-founded comparison for data augmentation methods.  
It contains 4,978 training utterances and 893 testing utterances. 
To simulate the data insufficient situations, we follow \citet{chen2016syntax,hou2018coling,shin2019vae},
and evaluate our model on two small proportions of the training data
which is \textit{small} proportion (1/40 of the original training set with 129 instances) and
\textit{medium} proportion (1/10 of the original training set with 515 instances).
We use a development set of 500 instances.

Snips \cite{Snips} dataset is collected from the  Snips personal voice assistant. 
There are 13,084 training utterances and 700 testing utterances.
We use another 700 utterances as the development set.
We also split the snips training set into \textit{small} proportion (1/100 of the original training set with 130 instances) and
\textit{medium} proportion (1/20 of the original training set with 654 instances).

\paragraph{Evaluation} Following previous works \cite{shin2019vae,hou2018coling}, 
we compute F1-score as evaluation metric with the \textit{conlleval} script.\footnote{\url{www.clips.uantwerpen.be/conll2000/ chunking/conlleval.txt}}

\paragraph{Implementation}
We built our Cluster2Cluster model with the transformer implemented by \citet{Wolf2019HuggingFacesTS}.
For pre-trained parameters, we used the GPT-2, which has 12 layers, 110M parameters and the hidden state dimension of 768. 
We used AdamW \cite{AdmaW} optimizer with initial learning rate 6.25e-5 or 5e-5 for training. 
We varied $\lambda$ in \{0.1, 0.02, 0.01, 0.002, 0.001\} and set $\gamma$ as 1.0.

Following previous works \cite{shin2019vae,hou2018coling}, 
we conduct experiments with Bi-LSTM as slot-filling model and train it with both original training data and data augmented by different data augmentation methods.
We use the same Bi-LSTM implements as previous work.\footnote{\url{github.com/AtmaHou/Bi-LSTM_PosTagger}}
The dimension of word embeddings and hidden states was set to 300 and 128, respectively.
We used GloVe \cite{pennington2014glove} to initialize word embedding.
We varied training batch size in \{16, 128\}, set dropout rate to 0.5, and trained the model with Adam as suggested by \citet{kingma2014adam}.

For all models, best hyperparameter settings are determined on the development set.
We report the average of 5 differently-seeded runs for each result. 


\subsection{Main Results for Data Augmentation}
Table \ref{tbl:main} shows the evaluation results of data augmentation methods on two slot filling datasets: ATIS and Snips.
To simulate data insufficient situations, we compare the proposed method with previous data augmentation methods with different proportions following previous works \cite{chen2016syntax,hou2018coling,shin2019vae}.
Baseline results are obtained with a Bi-LSTM slot-filling model trained on original training data. 
And results of each data augmentation methods are obtained with Bi-LSTM models that have the same architecture as the baseline but are trained with both original data and generated data.

On ATIS dataset, our model significantly outperforms the baseline model by 5.10 and 7.99 F-scores on medium and small proportion respectively. 
There are similar improvements on Snips dataset.
These improvements show the effectiveness of our augmentation method in the data-insufficient scenarios. 
When tested with data sufficient scenarios on full proportions, our model also brings improvements over baselines models. 
The improvements are narrowed comparing to those in data scarcity settings. 
We address this to the fact that full ATIS and Snips are large enough for slot-fillings, which limit the effects of additional synthetic data.
When we augment new data without generation pre-training, our performance drops but still achieves significant improvements in most settings, which shows the effectiveness of pre-training and C2C structure respectively.
We will discuss pre-training in detail later.

We compare our methods to two kinds of popular data augmentation methods for slot filling: rephrasing-based and sampling-based methods. 
Similar to our methods, rephrasing-based data augmentation methods reconstruct existing data into alternative expressions.\footnote{Traditional paraphrasing and back-translation methods are not compared here, because they are not capable to generate token-level annotation for sequence labeling problem.} 
For this kind of method, NoiseSeq2Seq \cite{DBLP:conf/interspeech/KurataXZ16} and Rel-Seq2Seq\cite{hou2018coling} 
learn seq2seq models to reconstruct the existing utterances.
To generate unseen expression, NoiseSeq2Seq Introduce noise to decoding, and Rel-Seq2Seq considering the relation between expression alternatives.
Slot Expansion \cite{shin2019vae} generates the new data by randomly replacing the slot values of existing utterances.
These methods argument each new utterance independently, thus often generate duplicated expressions that are helpless to improve slot-filling training. 
Our C2C model mitigates this by jointly encoding and decoding multiple utterances and considering the extensive relation between instances.
Such advantages result in higher diversity and help to achieve better performance.

For the second type of data augmentation, we compare with the sampling-based data augmentation method of C-VAE \cite{shin2019vae}.
C-VAE leverages a conditioned VAE model to sample new utterances and generates corresponding annotations at the same time.
It also faces the diversity problem, since it samples each new data independently. 
Our methods outperform this strong baseline on all the six slot-filling settings. 
The improvements come from the better diversity and fluency of the proposed Cluster2Cluster generation. 
Notably, we gain significant improvements of 9.63 and 3.35 F1-scores on Snips-small and ATIS-small. 
It shows that our methods are more effective in data scarcity situations.


\subsection{Analysis}
\begin{table}[t]
	\centering
	\footnotesize
	\begin{tabular}{lccc}
		\toprule
		\multirow{1}{*}{\textbf{Model}}
		& {\textbf{Full}} & {\textbf{Medium}} & {\textbf{Small}} \\	
		\midrule
		
		Ours & {\textbf{91.01}} & {\textbf{67.90}}& {\textbf{48.09}}  \\
		\quad - cluster-wise gen. & 90.28 & 66.23 & 45.93   \\
		\quad - diverse reg. & 90.32 & 66.11 & 44.32    \\
		\quad - dup. attention & 90.16 & 66.43 & 44.37    \\
		\bottomrule
	\end{tabular}
	\caption{\footnotesize
		The results of the ablation test.
	}\label{tbl:ablation}
\end{table}


\begin{table}[t]
	\centering
	\footnotesize
	\begin{tabular}{lccccc}
		\toprule
		\multirow{1}{*}{\textbf{Model}} & \multirow{1}{*}{\textbf{TF Layers}} 
		& {\textbf{Full}} & {\textbf{Medium}} & {\textbf{Small}} \\	
		\midrule
		
		w/ ~~pre-train & 12 & {\textbf{91.01}} & {\textbf{67.90}} & {\textbf{48.09}} \\
		w/o pre-train & 12 & 90.42 & 67.32 & 44.96 \\
		w/o pre-train & 2 & 90.33 & 67.49 & 46.94  \\
		w/o pre-train & 1 & 90.10 & 66.45 & 46.59  \\
		\bottomrule
	\end{tabular}
	\caption{\footnotesize
	Effect analysis of generation pre-training.
	}\label{tbl:pretrain}
\end{table}

\paragraph{Ablation Test}
We perform an ablation study to evaluate the importance of each component in C2C framework.
Table \ref{tbl:ablation} shows the results on Snips.
For the model without cluster-wise generation, 
we directly fine-tune GPT to generate new data in a seq-to-seq manner. 
The drops of F1-score demonstrate the superiority of the cluster-wise generation. 
If removing either Diverse-Oriented Regularization or Duplication-ware attention from the model, 
performance drops are witnessed. 
This shows that both of the two mechanisms help to improve slot-filling by encouraging diversity.





\paragraph{Effects of Generation Pre-training}
We analyze the impact of initializing C2C model with pre-trained language model. 
We randomly initialize C2C model and vary the model sizes to avoid overfitting caused by large model sizes.
As shown in Table \ref{tbl:pretrain}, the pre-training helps to improve the effects of data augmentation on all settings.
We attribute this to the fact that pre-training can improve generation fluency.
However, as revealed in both Table \ref{tbl:pretrain} and Table \ref{tbl:main}, the drops are limited compared to the overall improvements, which shows the inherent effectiveness of C2C model.




\begin{table}[t]
	\centering
	\footnotesize
	\setlength{\tabcolsep}{0.9mm}
	\resizebox{\linewidth}{!}{
	\begin{tabular}{lcccccc}
		\toprule
		\multirow{2}{*}{\textbf{Model}} &
				\multicolumn{3}{c}{\textbf{ATIS}} &
				\multicolumn{3}{c}{\textbf{Snips}} 	\\
				\cmidrule(lr){2-4}
				\cmidrule(lr){5-7}
				& {\textbf{Full}} & {\textbf{Medium}} & {\textbf{Small}} & {\textbf{Full}} & {\textbf{Medium}} & {\textbf{Small}} \\
		\midrule
		Baseline & 94.93 & 85.85 & 67.33 & 89.30 & 64.84 & 42.33\\
		~ + BERT & {\textbf{95.53}} & 91.18  &  82.56 
				& {\textbf{96.63}}  & 86.34 & 66.43 \\
		~ + BERT + Ours & 95.45 & {\textbf{92.13}} & {\textbf{85.92}} 
				&  96.12 & {\textbf{88.63}} & {\textbf{73.37}} \\	
		\bottomrule
	\end{tabular}
	}
	\caption{\footnotesize{Analysis of data augmentation effects with deep pre-trained embeddings}
	}\label{tbl:bert}
\end{table}


\paragraph{Effects over Deep Pre-trained Embeddings}
For data scarcity problem, deep pre-trained embeddings, such as BERT \cite{BERT}, are also demonstrated as an effective solution \cite{wang2020static}. 
To see whether DA is still effective when using deep pre-trained embeddings, we conduct DA experiments over a BERT-based slot-filling model.\footnote{We fine-tune the BERT-uncased-base model with AdamW optimizer and initial learning rate of 5e-5.}
As shown in Table \ref{tbl:bert}, although BERT greatly improves the performance of slot-filling, our model still achieved improvements on Medium and Small proportion data. 
This shows the effectiveness of our DA methods for data scarcity problems. 
Our augmentation method slightly lags the BERT-only model on Full proportion. 
We address this to the fact that full data is large enough for slot-filling and BERT can be misled by the noise within generated data.

%

\subsection{Evaluation for Generation Diversity}\label{sec:DED}

Increasing the diversity of generation is one of the essential goals of the data augmentation methods.
Following \citet{shin2019vae}, we evaluate the diversity of generated data from two aspects:
Inter and Intra. 
\texttt{Inter}: ratio of utterances that did not appear in the original training set. 
\texttt{Intra}: ratio of unique utterances among all generated new data.

Such metrics only measure the whole-sentence level diversity, but fail to measure expression  diversity at token level. 
To remedy this, we introduce a token-level diversity metric: Minimum Edit Distance (${\rm MED}$). 
For each generated utterance $\mathbf{u}'$, 
we calculate its ${\rm MED}$ to a set of utterances $C$ as ${\rm MED}(\mathbf{u}', C) = \min_{\mathbf{u} \in C}{\textsc{EditDistance}(\mathbf{u}, \mathbf{u'})}$.
${\rm MED}$ measures novelty of a sentence comparing to a set of existing sentences at token level. 
We report the averaged ${\rm MED}$ of each generated utterance to the original training set (\texttt{Inter}) and to the other generated utterances (\texttt{Intra}).

Table \ref{tbl:diverse} shows the evaluation of the generation diversity on the ATIS-Full. 
For Inter Diversity, our method significantly outperforms all previous methods
on both Ratio and average ${\rm MED}$ metrics.
We note that we can achieve the best diversity even evaluating the generated delexicalized utterances.
It shows the great ability of the C2C model in generating unseen expressions. 
This is mainly due to that cluster-wise encoding mechanism allows model to be aware of more existing expression during generation. 

For Intra Diversity, our method also achieves the best performances over the previous works.
These improvements show that considering relations between generated utterances can significantly reduce duplication.

\begin{figure}[t]
	\centering
	\footnotesize
	\begin{tikzpicture}
	\draw (0,0 ) node[inner sep=0] {\includegraphics[width=0.9\columnwidth, trim={0.28cm 1.1cm 8cm 2.9cm}, clip]{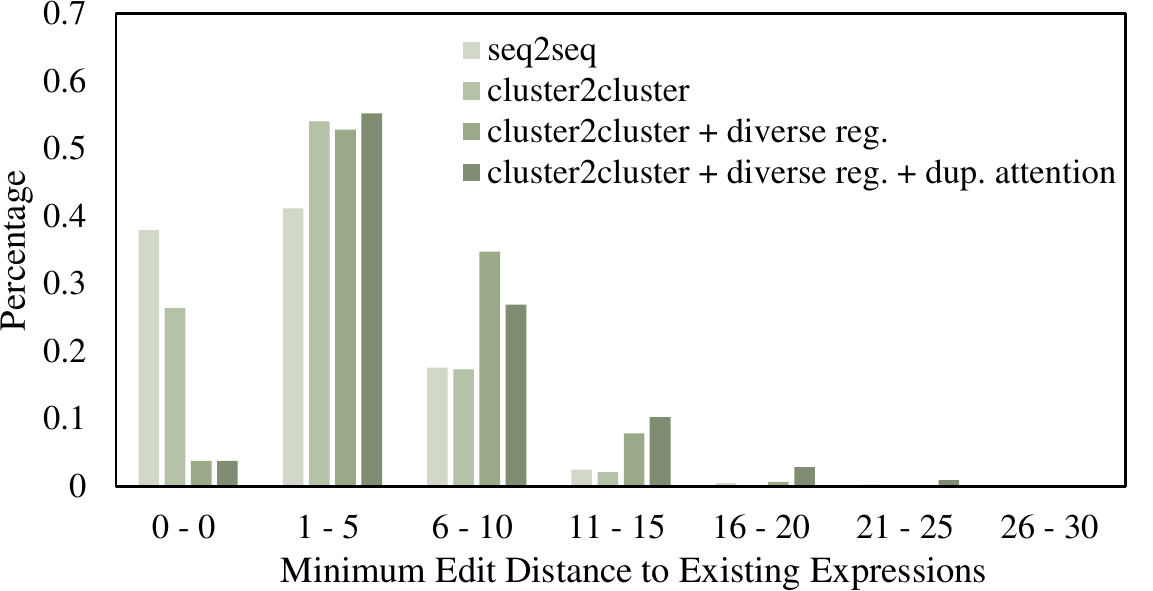}};
	\end{tikzpicture}
	\caption{\footnotesize
		Diversity distribution of generated expressions.  
	}\label{fig:diverse}
\end{figure}

\paragraph{Diversity Analysis}
To understand how the proposed method enhances expression diversity, we investigate the diversity distribution of generated delexicalized utterances on the ATIS-full.
We measure the diversity with Inter ${\rm MED}$.
As shown in Figure \ref{fig:diverse}, Seq2Seq generation yields more existing expressions, and the ${\rm MED}$ scores are mostly distributed in low-value areas.
Comparing to Seq2Seq, Cluster2Cluster model generally has higher ${\rm MED}$ scores.
This demonstrates the intrinsic advantage of the cluster-wise generation to generate new expressions.

When training the Cluster2Cluster model with Diverse-Oriented Regularization and Duplication-ware Attention, there is much fewer existing expressions within generated utterances, and we can see a continuous drifting of distribution towards higher diversity.
This shows that the proposed mechanisms help to generate more diverse utterances.

Also, we conduct case studies to see how C2C model generates unseen expressions (\textbf{See Appendix}). 
\begin{table}[t]
	\centering
	\footnotesize
	\renewcommand\arraystretch{1.1}
	\begin{tabular}{lcccc}
		\toprule
		\multirow{2}{*}{\textbf{Model}} &
		\multicolumn{2}{c}{\textbf{Inter}} &
		\multicolumn{2}{c}{\textbf{Intra}} 	\\
		\cmidrule(lr){2-3}
		\cmidrule(lr){4-5}
		& {\textbf{Ratio}} & {\textbf{MED}} & {\textbf{Ratio}} & {\textbf{MED}} \\	
		\midrule
		
		NoiseSeq2Seq & 74\% & 1.20 & 86\% & 1.03\\
		Rel-Seq2Seq & 96\% & 3.16 & 90\% & 1.81\\
		C-VAE & 23\% & 0.62  & 11\% & 0.42\\
		\midrule
		Ours (delexicalized) & {\textbf{96\%}} & {\textbf{5.88}} & {\textbf{92\%}} & {\textbf{5.55}}  \\
		Ours & {\textbf{100\%}} & {\textbf{9.03}} & {\textbf{95\%}} & {\textbf{4.85}}  \\
		\bottomrule
	\end{tabular}
	\footnotesize
	\caption{\footnotesize
		Diversity evaluation of utterance generation. 
	}\label{tbl:diverse}
\end{table}

\section{Related Work}
Data augmentation (DA) solves data scarcity problems by enlarging the size of training data \cite{fader-zettlemoyer-etzioni:2013:ACL2013,zhang2015character,daforslu,daforslu2,dafordst,li2019insufficient}.
Previous DA works propose back-translation methods \cite{backtranslate,backtranslate1} and paraphrasing methods \cite{paraphrase,paraphrase18,paraphrase19,gao2020paraphrase} to generate semantically similar sentences.
However, these DA methods are not applicable to the sequence labeling problem of slot-filling.
Because slot filling requires token-level annotations of semantic frame, while these methods can only provide sentence-level labels.

Spoken Language understanding, including slot filling and intent detection tasks, has drawn a lot of research attention recently
\cite{yao2013recurrent,7078572,DBLP:conf/interspeech/MesnilHDB13,DBLP:journals/taslp/MesnilDYBDHHHTY15,chen2016syntax,contextualslu,slusota2,slusota,slusota3}.
In this paper, we only focus on the slot filling task.
For data augmentation of slot filling, previous works focus on generation-based methods.
\citet{DBLP:conf/interspeech/KurataXZ16,hou2018coling,peng2020gpt} augment the training data with a Sequence-to-Sequence model. 
\citet{shin2019vae,yoo2019joint} introduced Variational Auto-Encoder \cite{Kingma2014vae} and jointly generate new utterances and predict the labels.
\citet{louvan2020simple} introduce simple rules to generate new utterances.
Different from our C2C framework, 
these methods augment each instance independently and often unconsciously generate duplicated expressions.

\section{Conclusion}
In this paper, we study the data augmentation problem for slot filling and propose a novel data augmentation framework C2C-GenDA, 
which generates new instances from existing training data in a cluster-to-cluster manner.
C2C-GenDA improves generation diversity by considering the relation between generated utterances and capturing more existing expressions. 
To further encourage diversity, we propose Duplication-aware Attention and Diverse-Oriented Regularization mechanism.
We introduce a Dispersed Cluster Pairing algorithm to construct cluster-to-cluster paraphrasing pairs for C2C-DA training.
Experiments show that the proposed framework can improve slot-filling by generating diverse new training data and outperform existing data augmentation systems of slot-filling.


\bibliography{AAAI21}

	\appendix
	
	\section*{Appendices}
	
	\section{Case Study of Generation}
	

	\begin{table*}[t]
		\centering
		\footnotesize
		\begin{tabular}{ll}\toprule
			\textbf{Type} & \textbf{$\mathbf{u}$ and $\mathbf{u}'$} \\
			\midrule
			\multirow{2}{*}{Replace Phrases} & $\mathbf{u}$: \textbf{show me the} flights from $<$from\_city$>$ to $<$to\_city$>$ \textbf{with stop in} $<$stop\_city$>$  \\
			& $\mathbf{u}'$: \textbf{give me} flights from $<$from\_city$>$ to$<$to\_city$>$ \textbf{with stopover in} $<$stop\_city$>$  \\  
			\midrule
			
			\multirow{2}{*}{Enrich Info} & $\mathbf{u}$: \textbf{i 'd like information on} all the flights from $<$from\_city$>$ to $<$to\_city$>$ on $<$depart\_date$>$   \\
			& $\mathbf{u}'$:\textbf{ i 'm sorry to see} all the flights \textbf{that i take} from $<$from\_city$>$ to $<$to\_city$>$ on $<$depart\_date$>$ \\      
			\midrule
			
			\multirow{2}{*}{Change Syntax} & $\mathbf{u}$: \textbf{how much is} a flight from $<$from\_city$>$ to $<$to\_city$>$   \\
			& $\mathbf{u}'$: \textbf{how much does} a flight \textbf{cost} from $<$from\_city$>$ to $<$to\_city$>$  \\      
			\midrule
			
			\multirow{2}{*}{Change Semantics} & $\mathbf{u}$: show \textbf{all airlines with flights} between $<$from\_city$>$ and $<$to\_city$>$ \\
			& $\mathbf{u}'$: show me \textbf{more airlines with seats} between $<$from\_city$>$ and $<$to\_city$>$  \\       
			
			\bottomrule
		\end{tabular}
		\caption{\footnotesize
			Case study of new expression. 
			For each generated example $\mathbf{u}'$, we find the most similar existing utterance $\mathbf{u}$ to it and compare the differences. 
		}\label{tbl:case_exp}
	\end{table*}
	
	We conduct case study to see how the proposed Cluster2Cluster model generates unseen expressions. 
	Specifically, we randomly picking a generated utterance $\mathbf{u}'$,
	and searching in the original training set $D$ for the most similar utterance $\mathbf{u} = \argmin_{\hat{\mathbf{u}} \in D}{\textsc{EditDistance}(\hat{\mathbf{u}}, \mathbf{u'})}$.
	To focus on expressions change, 
	we perform experiments on delexicalized utterances. 
	
	By comparing the difference between each pair of $\mathbf{u}'$ and $\mathbf{u}$, 
	we find the new expressions in several interesting manners: 
	Replace Phrases, Enrich Info, Change Syntax, Change Semantics. 
	The examples are listed in Table \ref{tbl:case_exp}. 
	We find that the most common new expressions are from replacing phrases and enriching information. 
	As shown from the given examples,
	Cluster2Cluster model can replace semantically similar phrases and bring additional information. 
	The phrase replacing ability is mainly learned from cluster-to-cluster training data, 
	and the enriched information often comes from the pretraining process. 
	As for Change Syntax, it is interesting to see that Cluster2Cluster can yield new expression by using alternative syntax, which is very close to humans in paraphrasing. 
	

\end{document}